\documentclass[runningheads]{llncs}
\usepackage[T1]{fontenc}
\usepackage{etoolbox} % For conditionals

\usepackage{times}
\usepackage{latexsym}
\usepackage{xspace, subfigure}
\usepackage{graphicx}
\usepackage{hyperref}
\usepackage{adjustbox}
\usepackage{array}
\usepackage{balance}
\usepackage{multirow}
\usepackage{babel, caption}
\usepackage{amsmath, amssymb, amsfonts}
\usepackage{booktabs}
\newcommand{\our}{\textsc{ColdSelect}\xspace}
\newcommand\numberthis{\addtocounter{equation}{1}\tag{\theequation}}
\DeclareMathOperator*{\argmax}{arg\,max}

\usepackage{multirow}
\usepackage{xparse}
\usepackage{footmisc}
\begin{document}
\title{Modeling Data Diversity for Joint Instance and Verbalizer Selection in Cold-Start Scenarios}

\author{Mohna Chakraborty\orcidID{0000-0003-3112-7445\thanks{The first two authors contributed equally to this work.}},
Adithya Kulkarni\orcidID{0000-0002-4625-4212}$^*$, \and
Qi Li\orcidID{0000-0002-3136-2157}}
\authorrunning{Mohna Adithya et al.}
% First names are abbreviated in the running head.
% If there are more than two authors, 'et al.' is used.
%
\institute{Iowa State University, Ames, Iowa, 50011, USA \\
\email{\{mohnac, aditkulk, qli\}@iastate.edu}}
\maketitle              % typeset the header of the contribution

\begin{abstract}
Prompt-based methods leverage the knowledge of pre-trained language models (PLMs) trained with a masked language modeling (MLM) objective; however, these methods are sensitive to template, verbalizer, and few-shot instance selection, particularly in cold-start settings with no labeled data. Existing studies overlook the dependency between instances and verbalizers, where instance-label probabilities depend on verbalizer token proximity in the embedding space. To address this, we propose \our, a joint verbalizer and instance selection approach that models data diversity. \our maps PLM vocabulary and $h_{[MASK]}$ embeddings into a shared space, applying dimensionality reduction and clustering to ensure efficient and diverse selection. By optimizing for minimal uncertainty and maximal diversity, \our captures data relationships effectively. Experiments on eight benchmarks demonstrate \our superiority in reducing uncertainty and enhancing generalization, outperforming baselines in verbalizer and few-shot instance selection for cold-start scenarios. 

\end{abstract}
\section{Introduction}
\label{introduction}
Pre-trained language models (PLMs) trained with the masked language modeling (MLM) objective~\cite{liu2019roberta} have become essential for various NLP downstream tasks~\cite{10.1145/3534678.3539386}, as their training on extensive corpora allows them to capture rich contextual information. Prompt-based methods capitalize on this by transforming classification tasks into cloze-style tasks~\cite{gao2021making}, where PLMs predict the [MASK] token using suitable vocabulary tokens. This alignment with the pre-training objective allows prompt-based methods to deliver strong performance, even with limited labeled data. In this study, we focus on moderately sized masked language models, as generative models pose challenges such as high deployment costs on local hardware and privacy concerns when using APIs~\cite{achiam2023gpt} for sensitive data. Our approach balances efficiency, performance, and data security.

Prompt-based methods rely on two key components, templates and verbalizers, that together unlock the potential of PLMs for downstream tasks. Templates reframe input data into cloze-style tasks, enabling the model to leverage its pre-trained MLM capabilities, while verbalizers map the model's vocabulary predictions to class labels, serving as the crucial link between token outputs and task-specific categories. Templates can be manually designed~\cite{schick2021exploiting,wang2022prompt}, automatically generated~\cite{gao2021making,liu2023gpt}, or constructed continuously~\cite{lester2021power,li2021prefix,liu2023global}. Similarly, verbalizers can be divided into three categories: manual~\cite{schick2021exploiting,schick2021s}, search-based~\cite{schick2020automatically,gao2021making,shin2020autoprompt}, and soft verbalizers~\cite{hambardzumyan2021warp,zhangdifferentiable}. 
While the manual creation of templates and verbalizers provides a straightforward approach, it is inherently limited by human interpretation, often resulting in suboptimal representations. In contrast, automatic and continuous methods reduce manual effort, dynamically adapting to optimize the model's performance. However, the effectiveness of prompt-based methods remains highly sensitive to the choice of templates~\cite{chakraborty2023zero}, verbalizers~\cite{gao2021making}, and few-shot labeled instances~\cite{yu2023cold}. This sensitivity underscores the need for approaches that better model the diversity and complexity of data distributions to ensure robust and generalizable performance. 

To enhance the performance of prompt-based methods, we focus on annotating instances and obtaining verbalizer tokens within a given labeling budget~\cite{pmlr-v216-kulkarni23a}. Efficient use of this budget requires a balanced approach to both instance and verbalizer selection, as these elements are interrelated. Ignoring this relationship can result in suboptimal outcomes. Existing methods for verbalizer selection~\cite{gao2021making,wang2024manifold} rely on randomly chosen few-shot instances, often lacking the diversity needed for robust generalization. Similarly, instance selection approaches~\cite{yu2023cold} using fixed, manually designed verbalizers fail to capture data variability or adapt to nuanced label distributions.
These studies overlook the dependency between instance and verbalizer selection. Under the MLM objective, an instance is more likely to predict a label accurately if the verbalizer token lies nearby in the embedding space. Ignoring this relationship results in redundant examples, noisy data, and outliers, which degrade generalization and robustness, especially in cold-start scenarios without labeled data.

To address the aforementioned challenges of data diversity and uncertainty in cold-start scenarios, we propose \our, a novel method that jointly selects verbalizers and few-shot instances by modeling data diversity. Modeling data diversity ensures the selection of diverse instances that represent the corpus comprehensively, reducing redundancy and improving generalization. For example, in sentiment analysis tasks, including instances with varying sentiment intensities helps the model learn nuanced distinctions, while in news classification, diverse examples across categories ensure balanced representation. At the same time, diverse verbalizer tokens for a class, such as mapping ``great'' and ``magnificent'' to \textit{``positive''} class, effectively capture label semantics and avoid oversimplified mappings. Jointly optimizing instance and verbalizer selection within a single labeling budget maximizes efficiency and minimizes noise and redundancy.

\our maps pre-softmax embeddings of PLM vocabulary tokens and $h_{[MASK]}$ embeddings into a shared space for efficient comparison. Dimensionality reduction using PCA~\cite{wold1987principal} enhances computational efficiency, while clustering methods like KMeans~\cite{macqueen1967some} and refinement with negative silhouette loss~\cite{rousseeuw1987silhouettes} ensure robust, well-separated clusters that capture the data's diversity.
Within the resulting clusters, instance and verbalizer selection are guided by an optimization framework that operates under a labeling budget $\mathcal{B}$. This framework minimizes labeling uncertainty at each step by balancing three critical factors: intra-cluster cohesion, which ensures selected instances are representative; inter-cluster separation, which avoids redundancy across clusters; and impurity, which captures label diversity. At each timestamp, \our identifies the most informative clusters, from which the instances to annotate and verbalizer tokens are selected. By integrating these steps, \our ensures that the selected tokens and instances reflect the dataset's diversity and maximize the model's generalization capability, ultimately improving performance in prompt-based tasks.

In summary, the contributions of this study are as follows:
\begin{enumerate}
    \item To the best of our knowledge, this is the first method to jointly and automatically select instances to annotate and verbalizer tokens in a cold-start setting. By modeling data diversity using shared embedding spaces, clustering techniques, and a novel selection-based optimization framework, our approach ensures robustness and generalization, effectively addressing sensitivity in prompt-based methods.
    \item The instance and verbalizer selection process is formulated as an optimization problem designed to minimize labeling uncertainty at each step of the selection process.
    \item Comprehensive experiments on benchmark datasets show that \our successfully models data diversity and the selected instances and verbalizer tokens reduce labeling uncertainty, leading to improved accuracy.
\end{enumerate}

\section{Related Works}
\label{related_works}
Despite the remarkable success of PLMs, their application in specific tasks remains challenging, particularly in cold-start scenarios where no labeled data is available. This limitation has led to the growing interest in prompt-based methods, which reformulate downstream tasks into cloze-style tasks to better align with the MLM pre-training objective. Prompt-based approaches rely on three key components: templates, verbalizers, and, optionally, a few labeled instances for fine-tuning. While significant progress has been made in automatic and continuous template generation~\cite{schick2021exploiting,wang2022prompt,liu2023gpt,gao2021making,lester2021power,li2021prefix,liu2023global}, the automatic selection of verbalizers and few labeled instances remains underexplored, particularly in cold-start settings. Below, we review related work in these two areas.

\noindent \textbf{Few-Shot Instance Selection in Cold-Start Settings.}
Selecting diverse and representative few-shot instances is essential for improving the performance of prompt-based methods, particularly in cold-start scenarios where only unlabeled data is available. Early approaches~\cite{muller2022active} relied on clustering and heuristic-driven selection, but their inability to account for inter-sample diversity limited their effectiveness. Subsequent methods~\cite{chang2021training,yuan2020cold} utilized PLMs, leveraging embedding spaces or MLM loss to guide instance selection. While these strategies were task-agnostic, they often struggled with misalignment between pre-training objectives and downstream tasks, leading to suboptimal results. Recent efforts~\cite{liu2022makes,su2022selective} have focused on few-shot selection for large-scale language models through in-context learning. However, these methods lack a cohesive framework to simultaneously address data diversity and labeling uncertainty, leaving significant room for improvement.

\noindent \textbf{Verbalizer Selection.}
Verbalizers play a crucial role in mapping model predictions to class labels. Early approaches~\cite{schick2021exploiting,schick2021s} relied on manually designed verbalizers, which, while effective, were time-consuming and susceptible to human bias. To automate this process, search-based methods~\cite{shin2020autoprompt,schick2020automatically} identified tokens that maximized conditional probabilities within the LLM vocabulary. However, these approaches often generated tokens that lacked contextual relevance. Enhancements using semantically similar tokens from external knowledge bases~\cite{hu2022knowledgeable} improved token quality but failed to address data diversity and struggled with scalability in large vocabularies and few-shot scenarios. Soft verbalizers~\cite{hambardzumyan2021warp,zhangdifferentiable} mitigated some of these limitations by learning continuous embeddings but required substantial labeled data, making them unsuitable for few-shot settings. More recently, prototypical verbalizers~\cite{cui2022prototypical} leveraged few-shot training data to generate prototype embeddings, achieving state-of-the-art performance in automated verbalizer design. However, even these methods often fell short of manual verbalizers in certain cases, highlighting the need for further improvement. More recently, \cite{wang2024manifold} proposed the tuning-free LLE-INC method, re-embedding the verbalizer space using intra-class neighborhood relationships to enhance the design.

Unlike previous studies, \our is the first approach to jointly select instances and verbalizers while explicitly modeling their interdependence and data diversity in cold-start settings. 

\section{Preliminaries}
\label{preliminaries}

\begin{figure}[t]
    \centering
    \includegraphics[width=\textwidth]{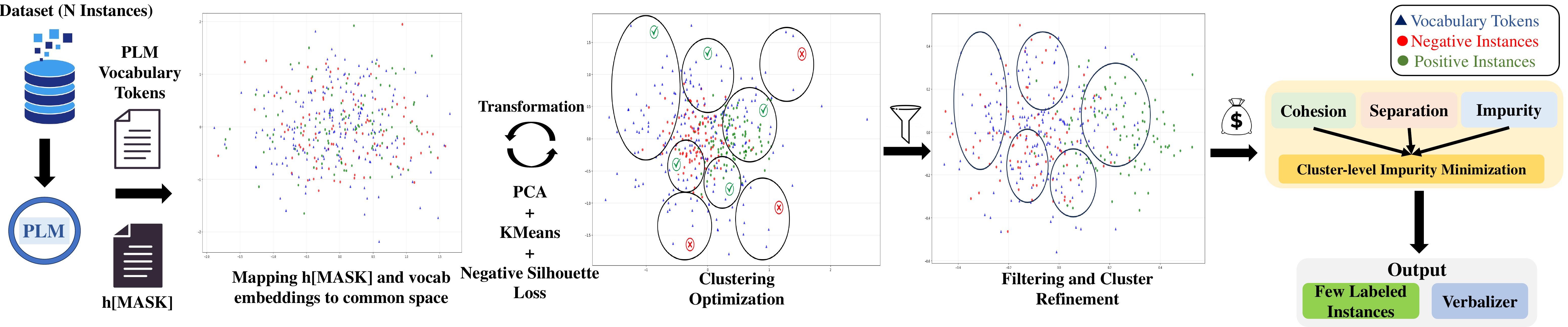}
    \caption{Overview of \our}
    \label{fig:intro_diagram}
    \vspace{-2em}
\end{figure}

In this section, we describe the process of obtaining prediction probabilities in prompt-based learning. Given a template $\mathcal{T}$, a verbalizer $\mathcal{M}:\mathcal{Y}\rightarrow\mathcal{V}$ that maps the class label space $\mathcal{Y}$ to tokens in the PLM vocabulary $\mathcal{V}$, and an input instance $\mathcal{I}$ from the unlabeled corpus $\mathcal{D}$, the probability of $\mathcal{I}$ being assigned a label $y \in \mathcal{Y}$ is defined as:
\begin{align*}
    p(y|\mathcal{I}) &= p([MASK] = \mathcal{M}(y) | \mathcal{I}_{\mathcal{T}}) = \frac{\exp(w_{\mathcal{M}(y)} \cdot h_{[MASK]})}{\sum_{y' \in \mathcal{Y}} \exp(w_{\mathcal{M}(y')} \cdot h_{[MASK]})}, \numberthis
    \label{eq1}
\end{align*}
where $\mathcal{I}_{\mathcal{T}} = \mathcal{T}(\mathcal{I})$ is the text obtained by applying the template $\mathcal{T}$ to the instance $\mathcal{I}$, resulting in a sentence with exactly one masked token ([MASK]). Here, $h_{[MASK]}$ represents the embedding of the [MASK] token, and $w_{v}$ is the pre-softmax token embedding for the token $v \in \mathcal{V}$ in the PLM's vocabulary. The predicted label for the instance $\mathcal{I}$ is the label $y \in \mathcal{Y}$ with the highest predicted probability.

\section{Methodology}
\label{methodology}
This section introduces \our, a method for jointly selecting verbalizers and few-shot instances by effectively modeling data diversity. Section \ref{problem formulation} outlines the problem, and Section \ref{selection_dependency} highlights the instance-verbalizer relationship in cold-start settings. Given a dataset $\mathcal{D}$ with $N$ instances, \our begins by extracting embeddings of the PLM's vocabulary tokens and $h_{[MASK]}$ embeddings of dataset instances, mapping them into a shared embedding space. To capture the dataset's diversity, \our applies PCA for dimensionality reduction, followed by KMeans clustering and negative silhouette loss to produce robust, well-separated clusters. To refine these clusters, vocabulary-only clusters are discarded, and instances from instance-only clusters are reassigned to the nearest mixed clusters, ensuring that all clusters contain both instances and vocabulary tokens. Section \ref{grouping} details the cluster creation process. The refined clusters are then passed to the Selection and Annotation module, which uses cohesion, separation, and impurity metrics to model cluster uncertainty and selects a subset of instances to obtain annotations and verbalizer tokens under the given budget $\mathcal{B}$. Section \ref{proposed} provides details about the module, and Section \ref{optimal_policy} demonstrates the optimality of the proposed selection process. Figure \ref{fig:intro_diagram} provides an overview of \our.

\subsection{Problem Formulation}
\label{problem formulation}

Given a PLM $\mathcal{L}$ trained with MLM objective, an unlabeled corpus $\mathcal{D}$ containing $N$ instances, a template $\mathcal{T}$, and a labeling budget $\mathcal{B}$, the objective is to minimize uncertainty in classifying instances in $\mathcal{D}$ into predefined labels $\mathcal{Y}$ (binary or multi-class). 

\subsection{Relationship between Instance and Verbalizer Selection}
\label{selection_dependency}
In prompt-based learning, the probability of assigning a label $y \in \mathcal{Y}$ to an input instance $\mathcal{I}$ is defined in Eq. (\ref{eq1}). In the equation, the dot product $w_{\mathcal{M}(y)} \cdot h_{[MASK]}$ is maximized when the embeddings are similar or have high cosine similarity, assuming normalized embeddings:
\begin{align}
    \text{cos\_sim}(w_{\mathcal{M}(y)}, h_{[MASK]}) = \frac{w_{\mathcal{M}(y)} \cdot h_{[MASK]}}{\|w_{\mathcal{M}(y)}\| \|h_{[MASK]}\|}.
\end{align}
To ensure optimal classification, the verbalizer token $w_{\mathcal{M}(y)}$ must be close to the $h_{[MASK]}$ embedding of instances assigned to label $y$ in the shared embedding space.\\

\subsection{Modeling Data Diversity for Cluster Creation}
\label{grouping}
Prompt-based learning utilizes PLMs to extract pre-softmax embeddings $w_v$ for vocabulary tokens $v \in \mathcal{V}$ and $h_{[MASK]}$ embeddings for instances in the dataset $\mathcal{D}$. Since both types of embeddings are derived from the same PLM, they are mapped into a shared embedding space to enable meaningful comparisons. However, the high dimensionality of these embeddings can result in uniformly high cosine similarity values, diminishing their discriminative power. To address this, we first perform dimensionality reduction to enhance the separability of embeddings. This is followed by clustering and cluster optimization, ensuring the effective modeling of data diversity.\\

\textbf{Dimensionality Reduction with PCA:}
To reduce the dimensionality of embeddings, we apply Principal Component Analysis (PCA)~\cite{wold1987principal}, which projects the embeddings into a lower-dimensional space while retaining most of the variance:
\begin{align}
    z = X W, \quad W = \argmax_W \|X W\|_F, \quad \text{s.t. } W^\top W = I,
\end{align}
where $X$ is the matrix of original embeddings, $W$ is the transformation matrix, and $z$ is the reduced embedding. After reduction, we normalize the embeddings to ensure cosine similarity effectively represents the dot product.\\

\textbf{Clustering with KMeans:}
To cluster the embeddings based on their similarity, we use KMeans clustering~\cite{macqueen1967some}, where the number of clusters is set to $K$. KMeans minimizes the within-cluster variance:
\begin{align}
    \mathcal{L}_{kmeans} = \sum_{k=1}^{K} \sum_{x \in C_k} \|x - \mu_k\|^2,
\end{align}
where $C_k$ is the set of points in cluster $k$, and $\mu_k$ is the cluster centroid. Clustering is performed to group vocab tokens and instance embeddings to maximize intra-cluster similarity and facilitate the subsequent selection of verbalizers and instances.\\

\textbf{Optimizing Clustering with Negative Silhouette Loss:}
Since KMeans is sensitive to initialization and may produce suboptimal clusters, we refine the clustering using negative silhouette loss~\cite{rousseeuw1987silhouettes}, which measures cluster cohesion and separation. The silhouette score for a point $i$ is defined as:
\begin{align}
    S(i) = \frac{b(i) - a(i)}{\max(a(i), b(i))},
\end{align}
where $a(i)$ is the average distance to other points in the same cluster, and $b(i)$ is the smallest average distance to points in any other cluster. The objective is to minimize the negative silhouette score:
\begin{align}
    \mathcal{L}_{sil} = - \frac{1}{N} \sum_{i=1}^{N} S(i),
\end{align}
where $N$ is the total number of points. This optimization ensures clusters are well-separated and cohesive, improving the reliability of the clustering process.\\

\textbf{Filtering and Cluster Refinement:}
After clustering, three types of clusters typically emerge: (1) \textit{Mixed Clusters:} Contain both token and instance embeddings, (2) \textit{Token-Only Clusters:} Contain only token embeddings, often representing outliers, and (3) \textit{Instance-Only Clusters:} Contain only instance embeddings, typically distant from token distributions.
We discard token-only clusters as outliers and reassign instances from instance-only clusters to the nearest mixed cluster based on cosine similarity to the centroid:
\begin{align}
    C_{assign} = \argmax_{C_k \in \mathcal{C}} \text{cos\_sim}(\mu_k, h_{\mathcal{I}}),
\end{align}
where $\mu_k$ is the centroid of cluster $C_k \in \mathcal{C}$, and $h_{\mathcal{I}}$ is the $h_{[MASK]}$ embedding of instance $\mathcal{I}$. This refinement ensures all clusters are meaningful and suitable for verbalizer and instance selection.

By combining dimensionality reduction, clustering, optimization, and refinement, our approach ensures that the final clusters capture the diversity and dependency between PLM vocab token and instance embeddings, laying the foundation for robust instances for annotation and verbalizer selection.

\subsection{Modeling Uncertainty for Cluster Selection and Annotation}
\label{proposed}
The objective of \our is to minimize uncertainty in classifying instances in $\mathcal{D}$ into pre-defined labels $\mathcal{Y}$ (binary or multi-class). To achieve this, we leverage three key factors, cohesion, separation, and impurity, that collectively model cluster uncertainty. This framework ensures efficient use of the given labeling budget $\mathcal{B}$ in a cold-start setting. Below, we outline the motivation and mathematical formulation for each step in \our.

    \textbf{Cohesion:} For dense clusters where embeddings are close to the cluster centroid $\mu_k$, the probability that all instances belong to the same class is high. Cohesion models cluster density as:
    \begin{align}
        cohesion(C_k) = \frac{1}{|C_k|} \sum_{x \in C_k} \text{cos\_sim}(x, \mu_k),
        \label{eq_cohesion}
    \end{align}
    where $x \in C_k$ are the embeddings in cluster $C_k$, and $\text{cos\_sim}(x, \mu_k)$ measures their similarity to the cluster centroid.

    \textbf{Separation:} Clusters far from others may represent outlier classes where all instances belong to the same class. Separation quantifies the distance between clusters as:
    \begin{align}
        separation(C_k) = \max_{C_{k'} \neq C_k \in \mathcal{C}} \text{cos\_sim}(\mu_k, \mu_{k'}),
        \label{eq_separation}
    \end{align}
    where $\mu_{k'}$ is the cluster centroid of cluster $C_{k'}$.

    \textbf{Impurity:} Dense clusters may still contain instances from multiple classes, while sparse clusters can have low label diversity. Impurity models label diversity as:
    \begin{align}
        impurity(C_k) = 1 - \frac{\max_{l \in \mathcal{L}} count_{C_k}(l)}{total(C_k)},
        \label{eq_impurity}
    \end{align}
    where $\mathcal{L}$ is the set of labels, $count_{C_k}(l)$ is the number of instances with label $l$ in cluster $C_k$, and $total(C_k)$ is the total number of instances in $C_k$.

\paragraph{Instance Classification Uncertainty Minimization}
We model the uncertainty minimization problem as a cluster-level impurity minimization task. Since every instance belongs to a cluster, reducing impurity at the cluster level effectively minimizes uncertainty in instance classification. To achieve this, we prioritize clusters with high impurity at each step, operating on the principle that annotating instances within these clusters will significantly reduce their impurity. Additionally, we incorporate intra-cluster cohesion and inter-cluster separation to ensure the selection of representative clusters while avoiding redundancy. As a result, at each step $T$, the cluster that maximizes the following equation is selected for annotation.
\begin{align}
    \mathcal{C}_{T} = \argmax_{C_k \in \mathcal{C}} \mathbb{E} \big[cohesion(C_k) + separation(C_k) + impurity(C_k)\big],
\end{align}
 The inclusion of $cohesion(C_k)$ in the cluster selection process inherently favors dense clusters, which is beneficial, particularly under low labeling budgets. Dense clusters, characterized by closely clustered embeddings, often indicate that instances belong to the same class, making them ideal for efficient labeling. Prioritizing these clusters ensures that each labeled instance has maximum impact, minimizing uncertainty while conserving resources. Additionally, this strategy reduces noise in the early stages, creating a strong foundation for subsequent labeling. Over time, the balance between cohesion, separation, and impurity ensures that sparse and diverse clusters are also addressed, leading to optimal resource allocation and improved model performance.

\paragraph{Initialization for Cold-Start Settings}
In a cold-start scenario, where no labeled instances are initially available, the metrics are initialized as follows: \textit{cohesion} is computed using Eq. (\ref{eq_cohesion}), \textit{separation} is determined using Eq. (\ref{eq_separation}), and \textit{impurity} is set to $0$ for all groups.

\paragraph{Dynamic Updates for Cohesion and Separation}
To incorporate instance labeling dynamically, we replace static cluster centroids with embeddings of selected verbalizer tokens. Verbalizer tokens provide a more contextually relevant reference for cluster evaluation, as class probabilities are determined by the dot product between the $h_{[MASK]}$ embedding of an instance and these tokens. Unlike fixed centroids, verbalizer tokens adapt as labels are assigned, capturing evolving cluster dynamics effectively. Eq. (\ref{eq_cohesion}) and Eq. (\ref{eq_separation}) are updated as follows:\\
    \textbf{Cohesion:}
    \begin{align}
        cohesion(C_k) = \frac{1}{|C_k|} \sum_{x \in C_k} \max_{v \in \mathcal{V} \& v \in C_k} \text{cos\_sim}(x, v),
    \end{align}
    where $v \in \mathcal{V}$ are verbalizer tokens in cluster $C_k$.\\
    \textbf{Separation:}
    \begin{align}
        separation(C_k) = \max_{v \in \mathcal{V} \& v \notin C_k} \text{cos\_sim}(\mu_k, v).
    \end{align}
\paragraph{Labeling Policy}
We categorize the selected cluster $\mathcal{C}_{T}$ as labeled if at least one instance in the cluster is already labeled. Otherwise, it is categorized as unlabeled. Depending on the categorization, the instance selection for labeling proceeds as follows:\\
    \noindent If $\mathcal{C}_{T}$ is \textbf{unlabeled}, select the instance nearest to the cluster centroid $\mu$ for labeling:
    \begin{align}
        \mathcal{I}_{select_{T}} = \argmax_{x \in \mathcal{C}_{T}} \text{cos\_sim}(x, \mu).
    \end{align}
    Assign the label obtained for $\mathcal{I}_{select_{T}}$ to the nearest vocab token and add it to $\mathcal{V}$:
    \begin{align}
        v_{select_T} = \argmax_{v \in \mathcal{C}_{T}} \text{cos\_sim}(v, \mathcal{I}_{select_T}).
        \label{vocab}
    \end{align}
    \noindent If $\mathcal{C}_{T}$ is \textbf{labeled}, select the instance farthest from already labeled instances and obtain verbalizer token following Eq. (\ref{vocab}):
    \begin{align}
        \mathcal{I}_{select_T} = \argmax_{x \in \mathcal{C}_{T}} \min_{x \neq x' \in \text{labeled}} \text{cos\_sim}(x, x').
    \end{align}

\paragraph{Stopping Criterion}
The selection process continues until the labeling budget $\mathcal{B}$ is exhausted. By iteratively targeting clusters with maximum impurity, \our optimally selects instances and verbalizer tokens to reduce classification uncertainty and enhance model performance.

\subsection{Optimal Selection Process}
\label{optimal_policy}
The proposed approach reduces classification uncertainty in $\mathcal{D}$ by integrating \textbf{cohesion}, \textbf{separation}, and \textbf{impurity} metrics into the cluster selection process. This balanced scoring function enables effective exploration and exploitation, focusing on dense, distinct clusters initially and gradually addressing sparse or diverse clusters as labels are acquired. Metrics are initialized based on embedding proximity and dynamically updated during labeling, allowing adaptability in cold-start settings. By combining these metrics, the approach optimally utilizes the labeling budget $\mathcal{B}$ to reduce uncertainty, avoid redundancy, and ensure robust classification for diverse datasets.
\section{Experiments}
\label{experiments}

In this section, we evaluate the performance of \our in reducing uncertainty in classifying instances on several benchmark datasets from diverse domains: SST-2~\cite{socher2013recursive}, MR~\cite{pang2002thumbs}, CR~\cite{hu2004mining}, Subj~\cite{pang2004sentimental}, CoLA~\cite{warstadt2019neural}, AG News~\cite{zhang2015character}, Yelp~\cite{meng2019weakly}, and IMDB~\cite{maas2011learning}\footnote{The code is available at \url{https://github.com/Mohna0310/COLDSELECT}}. Table~\ref{table: Datasets Statistics} provides a summary of the datasets, including their type, number of classes, and the templates used. 

\begin{table}[t]
\centering
\caption{Statistics of the Datasets}
\resizebox{\columnwidth}{!}{
\begin{tabular}{c|c|c|c|c|c}
\hline
\textbf{Dataset} & \textbf{Type} & \textbf{|y|}   & \textbf{Labels}    & \textbf{\#Test Instances} & \textbf{Template} \\ \hline
SST-2 & Sentiment Analysis & 2 & positive, negative & 872 & <S>. It was [MASK]. \\ \hline
MR & Sentiment Analysis & 2 & positive, negative & 2,000 & <S>. It was [MASK]. \\ \hline
CR & Sentiment Analysis & 2 & positive, negative & 2,000 & <S>. It was [MASK]. \\ \hline
Subj & Subjectivity Classification & 2 & subjective, objective & 2,000 & <S>. It was [MASK]. \\ \hline
CoLA & Acceptability Classification & 2 & grammatical, not grammatical & 1,042 & <S>. This is [MASK]. \\ \hline
AG News & News Classification & 4 & world, sports, business, technology & 7,600 & [MASK] News: <S>. \\ \hline
Yelp-full & Sentiment Analysis & 5 & very positive, positive, neutral, negative, very negative & 38,000 & <S>. It was [MASK]. \\ \hline
IMDB & Sentiment Analysis & 2 & positive, negative & 25,000 & <S>. It was [MASK]. \\ \hline
\end{tabular}
}
\label{table: Datasets Statistics}
\vspace{-2em}
\end{table}

\begin{table*}[t]
\caption{Few-shot instance selection results on three datasets using RoBERTa-base with standard finetuning, following \cite{yu2023cold}. Accuracy is reported on the test set, with best and runner-up models highlighted in bold and underlined, respectively.}
\resizebox{\columnwidth}{!}{%
\begin{tabular}{c|c|c|c|c|c|c|c|c|c|c|c|c|c}
\centering
\textbf{Dataset} & \textbf{|y|} & \textbf{|$\mathcal{B}$|} & \textbf{Random} & \textbf{Uncertainity}  & \textbf{CAL} & \textbf{BERT-KM} & \textbf{Coreset} & \textbf{Margin-KM} & \textbf{ALPS} & \textbf{TPC} & \textbf{PATRON} & \textbf{Random-g} & \textbf{\our}\\
\hline
\multirow{3}{*}{IMDB} & \multirow{3}{*}{2} & 32 & 80.2 & 81.9 & 77.8 & 79.2 & 74.5 & 76.7 & 82.2 & 82.8 & \underline{85.5} & 84.1 & \textbf{87.37} \\
 &  & 64 & 82.6 & 84.7 & 81.2 & 84.9 & 82.8 & 84.0 & 86.1 & 84.0 & \underline{87.3} & 85.3 & \textbf{88.39} \\
 &  & 128 & 86.6 & 87.1 & 87.9 & 88.5 & 87.8 & 88.2 & 87.5 & 88.1 & \underline{89.6} & 88.7 & \textbf{90.61} \\ \hline
\multirow{3}{*}{Yelp-full} & \multirow{3}{*}{5} & 32 & 30.2 & 32.7 & 36.6 & 35.2 & 32.9 & 32.7 & \underline{36.8} & 32.6 & 35.9 & 34.1 & \textbf{39.58} \\
 &  & 64 & 42.5 & 36.8 & 41.2 & 39.3 & 39.9 & 39.8 & 40.3 & 39.7 & \underline{44.4} & 41.5 & \textbf{46.72} \\
 &  & 128 & 47.7 & 41.3 & 45.7 & 46.4 & 49.4 & 47.1 & 45.1 & 46.8 & \underline{51.2} & 48.9 & \textbf{54.16} \\ \hline
\multirow{3}{*}{AG News} & \multirow{3}{*}{4} & 32 & 73.7 & 73.7 &  69.4 & 79.1 & 78.6 & 75.1 & 78.4 & 80.7 & \underline{83.2} & 81.2 & \textbf{84.69} \\
 &  & 64 & 80.0 & 80.0 & 78.5 & 82.4 & 82.0 & 81.1 & 82.6 & 83.0 & \underline{85.3} & 83.8 & \textbf{87.16} \\
 &  & 128 & 84.5 & 82.5 & 81.3 & 85.6 & 85.2 & 85.7 & 84.3 & 85.7 & \underline{87.0} & 86.1 & \textbf{88.26} \\ \hline
\end{tabular}
}
\label{table: Results4_1}
\vspace{-2em}
\end{table*}

\begin{table*}[t]
\caption{Few-shot instance selection results on three benchmark datasets using RoBERTa-base, following \cite{yu2023cold}. Prompt-based finetuning is performed, with accuracy reported on the test set. Best and runner-up models are highlighted in bold and underlined, respectively.}
\resizebox{\columnwidth}{!}{%
\begin{tabular}{c|c|c|c|c|c|c|c|c|c|c|c|c|c}
\centering
\textbf{Dataset} & \textbf{|y|} & \textbf{|$\mathcal{B}$|} & \textbf{Random} & \textbf{Uncertainity}  & \textbf{CAL} & \textbf{BERT-KM} & \textbf{Coreset} & \textbf{Margin-KM} & \textbf{ALPS} & \textbf{TPC} & \textbf{PATRON} & \textbf{Random-g} & \textbf{\our}\\
\hline
\multirow{3}{*}{IMDB} & \multirow{3}{*}{2} & 32 & 81.8 & 82.4 & 79.6 & 81.7 & 85.5 & 86.0 & 83.5 & 84.5 & \underline{86.5} & 85.2 & \textbf{89.48}\\
 &  & 64 & 85.6 & 86.0 & 81.1 & 84.2 & 87.8 & 87.6 & 84.4 & 85.8 & \underline{88.8} & 87.2 & \textbf{91.42} \\
 &  & 128 & 87.7 & 88.4 & 83.0 & 88.5 & 88.9 & 89.1 & 88.9 & 88.0 & \underline{89.3} & 88.5 & \textbf{91.23} \\ \hline
\multirow{3}{*}{Yelp-full} & \multirow{3}{*}{5} & 32 & 48.9 & 46.6 & 47.9 & 45.5 & 46.0 & 47.5 & 47.0 & 49.8 & \underline{50.5} & 49.1 & \textbf{53.31} \\
 &  & 64 & 51.0 & 49.9 & 49.4 & 51.9 & 48.8 & 52.6 & 52.8 & 52.3 & \underline{53.6} & 51.5 & \textbf{56.29}\\
 &  & 128 & 51.3 & 50.8 & 48.7 & 51.5 & 53.7 & 54.2 & 51.7 & 51.0 & \underline{55.6} & 53.2 & \textbf{57.16}\\ \hline
\multirow{3}{*}{AG News} & \multirow{3}{*}{4} & 32 & 83.1 & 82.8 & 81.4 & 84.9 & 85.1 & 84.6 & 84.2 & 85.6 & \underline{86.8} & 85.7 & \textbf{87.23} \\
 &  & 64 & 84.5 & 84.3 & 82.6 & 86.5 & 86.4 & 85.9 & 86.2 & 85.6 & \underline{87.4} & 86.8 & \textbf{88.18}\\
 &  & 128 & 84.9 & 83.1 & 83.0 & 87.6 & 87.5 & 87.1 & 87.5 & 87.0 & \underline{87.8} & 87.4 & \textbf{89.95} \\ \hline
\end{tabular}
}
\label{table: Results4}
\vspace{-2em}
\end{table*}

\begin{table*}[t]
\centering
\caption{Performance comparison of \our with Random, Random-g, and other methods on five benchmark datasets using RoBERTa-large LLM, following \cite{gao2021making}. Experiments use fixed manual templates with $K=16$ few-shot instances per class to obtain automatic verbalizers. The maximum budget $\mathcal{B}$ required by Random, Random-g, and \our are $66$, $52$, and $44$, respectively, showing \our's efficiency in balancing class labels and optimizing the labeling budget. Accuracy (\%) is reported on test set, with the best results highlighted in bold.}
\resizebox{\textwidth}{!}{%
\begin{tabular}{l|ccc|ccc|ccc|ccc}
\hline
\multirow{2}{*}{\textbf{Dataset}} & \multicolumn{3}{c|}{\textbf{Fine-tuning}} & \multicolumn{3}{c|}{\textbf{Prompt-based FT}} & \multicolumn{3}{c|}{\textbf{LM-BFF}} & \multicolumn{3}{c}{\textbf{ProtoVerb}} \\ \cline{2-13}
 & \textbf{Rand} & \textbf{Rand-g} & \textbf{\our} & \textbf{Rand} & \textbf{Rand-g} & \textbf{\our} & \textbf{Rand} & \textbf{Rand-g} & \textbf{\our} & \textbf{Rand} & \textbf{Rand-g} & \textbf{\our} \\ \hline
\textbf{SST-2} & 81.4 & 83.0 & \textbf{87.5} & 92.7 & 93.1 & \textbf{93.0} & 92.6 & 92.8 & \textbf{93.2} & 86.9 & 87.0 & \textbf{87.1} \\ 
\textbf{MR} & 76.9 & 77.8 & \textbf{79.5} & 87.0 & 87.5 & \textbf{88.9} & 86.6 & 87.0 & \textbf{87.8} & 60.0 & 62.4 & \textbf{66.2} \\ 
\textbf{CR} & 75.8 & 76.3 & \textbf{78.0} & 90.3 & 90.7 & \textbf{91.5} & 90.2 & 90.6 & \textbf{91.7} & 68.7 & 70.1 & \textbf{76.8} \\ 
\textbf{Subj} & 90.8 & 91.2 & \textbf{93.2} & 91.2 & \textbf{91.5} & 90.2 & 92.3 & 92.5 & \textbf{93.7} & 75.7 & 76.0 & \textbf{76.3} \\ 
\textbf{CoLA} & 72.4 & 73.2 & \textbf{73.8} & 52.9 & 56.2 & \textbf{58.3} & 52.9 & 53.6 & \textbf{54.5} & 53.9 & 55.3 & \textbf{58.9} \\ 
\hline
\end{tabular}
}
\label{table: Results5}
\vspace{-2em}
\end{table*}

\subsection{Evaluation Metrics}
We use \textbf{Accuracy (Acc.)} as the primary evaluation metric across all datasets to measure the effectiveness of \our and baselines in reducing classification uncertainty.

\subsection{Baseline Methods}
Since \our is the first approach to jointly select both verbalizers and few-shot instances, we compare it with baselines for (1) few-shot instance selection and (2) verbalizer selection.

\textit{Few-Shot Instance Selection Baselines:} \textbf{Random}: randomly selects samples for annotation. \textbf{Uncertainty}~\cite{schroder2022revisiting}: selects instances with the highest uncertainty post-calibration using entropy~\cite{lewis1994sequential}.
\textbf{CAL}~\cite{margatina2021active}: uses Kullback-Leibler (KL) divergence to guide sample selection.
\textbf{Coreset}~\cite{sener2018active}: minimizes the maximum Euclidean distance between a sample and its nearest cluster centroid.
\textbf{BERT-KM}~\cite{chang2021training}: clusters embeddings using KMeans and selects samples closest to centroids.
\textbf{Margin-KM}~\cite{muller2022active}: selects samples based on the margin between the two highest probabilities within clusters.
\textbf{ALPS}~\cite{yuan2020cold}: uses BERT’s MLM loss to compute surprisal embeddings for sample selection.
\textbf{TPC}~\cite{hacohen2022active}: selects instances with the highest density in each cluster.
\textbf{PATRON}~\cite{yu2023cold}: employs a partition-then-rewrite strategy to enhance sample diversity.
\textbf{Random-g}: randomly selects refined clusters at each step, bypassing the proposed selection process while adhering to the labeling policy.

\textit{Automatic Verbalizer Selection Baselines:} \textbf{LM-BFF}~\cite{gao2021making}: uses T5 to automatically generate verbalizers with few-shot examples.
\textbf{ProtoVerb}~\cite{cui2022prototypical}: constructs prototype-based verbalizers using contrastive learning. For verbalizer selection, we also compare \our against manual verbalizers created by humans. 

\subsection{Experimental Settings}
We conduct experiments using RoBERTa-base (125M parameters) and RoBERTa-large (355M parameters) PLMs, following the setups of~\cite{yu2023cold} and~\cite{gao2021making}, respectively. For KMeans clustering, we set the random seed to 42 and the number of clusters to 40 and performed cluster optimization over five iterations.

\begin{table*}[t]
\centering
\caption{Ablation study on \our's performance with $\mathcal{B}=32$ using RoBERTa-base and standard fine-tuning, following \cite{yu2023cold}. Accuracy (\%) on the test set is reported, with best and runner-up results highlighted in bold and underlined.}
\resizebox{\textwidth}{!}{%
\begin{tabular}{l|c|c|c|c|c|c}
\hline
\textbf{Model Variant}      & \textbf{Cohesion} & \textbf{Separation} & \textbf{Impurity} & \textbf{IMDB} & \textbf{AG News} & \textbf{Yelp-full} \\ \hline
Only Cohesion               & \checkmark        &                     &                   & 80.50                      & 78.30                         & 30.80                           \\ \hline
Cohesion + Separation       & \checkmark        & \checkmark          &                   & \underline{82.20}          & \underline{80.10}             & \underline{33.70}               \\ \hline
Cohesion + Impurity         & \checkmark        &                     & \checkmark        & 81.40                      & 79.00                         & 32.90                           \\ \hline
Cohesion + Separation + Impurity (\our) & \checkmark        & \checkmark          & \checkmark        & \textbf{87.37}             & \textbf{84.69}                & \textbf{39.58}                  \\ \hline
\end{tabular}
}
\label{table:ablation}
\vspace{-2em}
\end{table*}

\subsection{Results and Discussion}

The results in Tables \ref{table: Results4_1} and \ref{table: Results4} demonstrate \our's superior performance over PATRON and other baselines by effectively leveraging data diversity and label uncertainty. Unlike PATRON, which relies on prompt-based uncertainty propagation and a static partition-then-rewrite (PTR) strategy, \our dynamically updates cluster centroids with verbalizer token embeddings, ensuring better alignment with evolving label distributions and reducing noise from outliers. By integrating cohesion, separation, and impurity metrics, \our balances exploration and exploitation, enabling early-stage labeling of dense clusters while progressively addressing sparse or diverse ones. This adaptability allows \our to surpass PATRON across various datasets, including IMDB in Table \ref{table: Results4_1}, where it achieves 87.37\% accuracy at \(|\mathcal{B}|=32\) (1.9\% higher than PATRON), Yelp-full in Table \ref{table: Results4_1}, where it achieves largest accuracy gain over PATRON, indicating its robustness in handling highly imbalanced class distributions, and AG News in Table \ref{table: Results4}, where it scales effectively to 89.95\% at \(|\mathcal{B}|=128\). These results highlight \our's ability to optimize labeling budgets and handle complex, multi-class distributions more effectively than existing methods.

The results in Table \ref{table: Results5} demonstrate \our's effectiveness across all datasets and experimental setups, consistently outperforming Random and Random-g. For instance, in the Fine-tuning setting, \our achieves 87.5\% on SST-2, surpassing Random (81.4\%) and Random-g (83.0\%). Similarly, on MR and CR, \our outperforms both baselines, achieving 79.5\% and 78.0\%, respectively. In the Prompt-based FT setup, \our achieves 91.5\% on CR and 58.3\% on CoLA, significantly higher than Random and Random-g.
\our’s reduced labeling budget ($\mathcal{B}=44$) compared to Random ($\mathcal{B}=66$) and Random-g ($\mathcal{B}=52$) highlights its efficiency in selecting diverse, representative instances. By ensuring that each labeled instance maximally reduces uncertainty, \our optimally utilizes limited annotation budgets, validating its ability to optimize instance selection while improving performance across binary classification tasks.
\our shows greater improvements in datasets with complex class distributions, such as Yelp-full and AG News, compared to binary tasks like SST-2. The results suggest that clustering-based selection of \our is particularly effective in capturing fine-grained distinctions between similar classes.

The ablation study in Table \ref{table:ablation} further supports this by showing that removing impurity modeling leads to a drop in accuracy across IMDB, AG News, and Yelp-full datasets, reinforcing the importance of selecting diverse and representative instances. Using cohesion alone provides a baseline improvement by prioritizing dense clusters (e.g., 80.50\% on IMDB), but it lacks robustness. Adding separation enhances performance (e.g., 82.20\% on IMDB) by ensuring inter-cluster distinctiveness, reducing redundancy, and better capturing outlier classes. Including impurity with cohesion offers slight gains (e.g., 81.40\% on IMDB) by addressing label diversity within clusters. However, the full combination of all three metrics yields the highest accuracy across datasets (e.g., 87.37\% on IMDB), demonstrating their complementary roles in capturing cluster density, distinctiveness, and label diversity to optimize labeling budgets effectively.

\section{Conclusion}
\label{conclusion}

In this study, we proposed \our, a novel approach, for instance, and verbalizer selection in prompt-based learning, explicitly modeling data diversity and label uncertainty. By integrating cohesion, separation, and impurity metrics, \our effectively identifies representative instances and optimizes labeling budgets. Extensive experiments on benchmark datasets demonstrate that \our consistently outperforms state-of-the-art methods, achieving robust performance in both standard and prompt-based fine-tuning. These results validate the importance of modeling intra-cluster density, inter-cluster distinctiveness, and label diversity, making \our a powerful solution for challenging cold-start scenarios and enhancing generalization in prompt-based classification tasks.

The effectiveness of \our in optimizing annotation budgets makes it highly applicable in low-resource NLP settings, active learning scenarios, and domain adaptation tasks. By reducing redundancy in labeled instances and improving generalization, this method can enhance model performance in real-world applications such as customer sentiment analysis, fake news detection, and biomedical text classification. Moreover, its ability to optimize instance selection without extensive labeled data makes it particularly relevant for emerging fields like legal text classification, financial risk assessment, and cross-lingual NLP, where annotated data is often scarce. Future extensions of \our could explore its adaptability to multilingual and multimodal datasets, further broadening its impact across diverse AI applications.

\section{Limitations and Future Work}
While \our excels in data-diverse scenarios, its reliance on PCA-based dimensionality reduction may introduce biases, as it assumes linear separability. Future work could explore adaptive non-linear transformations like kernel PCA or deep representation learning to enhance cluster separability. Additionally, the fixed labeling budget may not suit all datasets; a dynamic allocation strategy based on entropy minimization and cluster impurity could improve efficiency.

\our also depends on KMeans clustering, which may not always capture complex structures in high-dimensional spaces. Alternative clustering methods like hierarchical clustering or DBSCAN could enhance robustness. Additionally, integrating reinforcement learning could further refine instance selection by dynamically adapting to dataset characteristics.
\section{Acknowledgement}
The work is supported by the US National Science Foundation under grant NSF-CAREER 2237831.

\bibliographystyle{splncs04}
\bibliography{references}

\end{document}